%% file: 0082.tex
\documentclass[runningheads]{llncs}
\usepackage{amsmath,amssymb}
\usepackage{graphicx}
\usepackage{bm}
\usepackage{mathrsfs}
\usepackage{amssymb}
\usepackage{cite}
\usepackage{color}
\usepackage{xcolor}
\usepackage{subfigure}
\usepackage{tabu}
\usepackage{bm}
\usepackage{xcolor}
\usepackage{threeparttable}
\usepackage{booktabs}
\usepackage[pagebackref=true,breaklinks=true,letterpaper=true,colorlinks,citecolor=blue,linkcolor=blue,bookmarks=false]{hyperref}

\begin{document}
%===========================================================

\title{Guided Feature Selection for Deep Visual Odometry \thanks{Supported by the National Key Research and Development Program of China (2017YFB1002601) and National Natural Science Foundation of China (61632003, 61771026).}} % Replace your paper's title here
%\titlerunning{Short paper title} % Replace an abstracted version of your paper's title here

%===========================================================

\author{Fei Xue\inst{1,4}, Qiuyuan Wang\inst{1,4}, Xin Wang\inst{1,4}, Wei Dong\inst{2} \\ Junqiu Wang\inst{3} \and Hongbin Zha\inst{1,4}}
%
%Please include author names in full in the paper, 
%If any authors have names that can be parsed into FirstName LastName in multiple ways, please include the correct parsing, in a comment to the volume editors:
%\index{Lastnames, Firstnames}

\authorrunning{F. Xue et al.} % A shorter version of authors' name
% First names are abbreviated in the running head.
% If there are more than two authors, 'et al.' is used.

%===========================================================

\institute{Key Laboratory of Machine Perception (MOE), School of EECS, Peking University \and Robotics Institute, Carnegie Mellon University \and Beijing Changcheng Aviation Measurement and Control Institute \and Cooperative Medianet Innovation Center, Shanghai Jiao Tong University \\
\email{\{feixue, wangqiuyuan, xinwang\_cis\}@pku.edu.cn} , \email{weidong@andrew.cmu.edu}\\ 
\email{jerywangjq@foxmail.com},
\email{zha@cis.pku.edu.cn}}

\maketitle

%===========================================================
\begin{abstract}
We present a novel end-to-end visual odometry architecture with guided feature selection based on deep convolutional recurrent neural networks. Different from current monocular visual odometry methods, our approach is established on the intuition that features contribute discriminately to different motion patterns. Specifically, we propose a dual-branch recurrent network to learn the rotation and translation separately by leveraging current Convolutional Neural Network (CNN) for feature representation and Recurrent Neural Network (RNN) for image sequence reasoning. To enhance the ability of feature selection, we further introduce an effective context-aware guidance mechanism to force each branch to distill related information for specific motion pattern explicitly. Experiments demonstrate that on the prevalent KITTI and ICL\_NUIM benchmarks, our method outperforms current state-of-the-art model- and learning-based methods for both decoupled and joint camera pose recovery.

\end{abstract}
%===========================================================
\section{Introduction}
\label{introduction}
Visual Odometry (VO) and Visual Simultaneous Localization and Mapping (V-SLAM) estimate camera poses from image sequences by exploiting the consistency between neighboring frames. As an essential task in computer vision, VO has been widely used in autonomous driving, robotics and augmented reality. Features play a key role in building consistency across images, and have been widely used in current VO/SLAM algorithms \cite{newcombe2011dtam, geiger2011stereoscan, mur2017orb-slam2}. Despite the success of these methods, they ignore the discriminative contributions of features to different motions. However, if specific motions, especially rotations and translations, can be recovered by related features, the problems of scale-drifting and error accumulation in VO can be mitigated.

Unfortunately, how to detect appropriate features for recovering specific motions remains a challenging problem. Handcrafted feature descriptors such as SIFT \cite{lowe2004sift} ORB \cite{orb2011}, \textit{etc}. are designed for general visual tasks, lacking the response to motions. Instead, geometry priors such as vanishing points \cite{lee2015real}, planar structures \cite{kim2017visual, straub2015real}, and depth of pixels \cite{kaess2009flow, tardif2008monocular, paz2008large} are used in VO algorithms for camera pose decoupling. These methods provide promising performance in certain environments. However, they have limited generalization ability and may suffer from noisy input. 

Rather than handcrafted features, Convolutional Neural Networks (CNNs) are able to extract deep features, which can encode high level priors and can be fed into Recurrent Neural Networks (RNNs) for end-to-end image sequences modeling and camera motion estimation. A few methods based on regular Long Short-Term Memory (LSTM) \cite{hochreiter1997lstm} have been proposed for camera motion recovery, such as DeepVO \cite{wang2017deepvo} and ESP-VO \cite{wang2017espvo}. While achieving promising performances, they did not take into account the different responses of visual cues to motions, thus may output trajectories with large error.

In this paper, we aim to explore the possibility to select features with high discriminative ability for specific motions. Therefore, we can relax the assumptions of scenes required in previous works. We present a novel context-aware recurrent network that learns decoupled camera poses using selected features, as shown in Fig.~\ref{fig:structure}. The main contributions include:

\begin{figure}[t]
	\centering
	\includegraphics[height=7.cm]{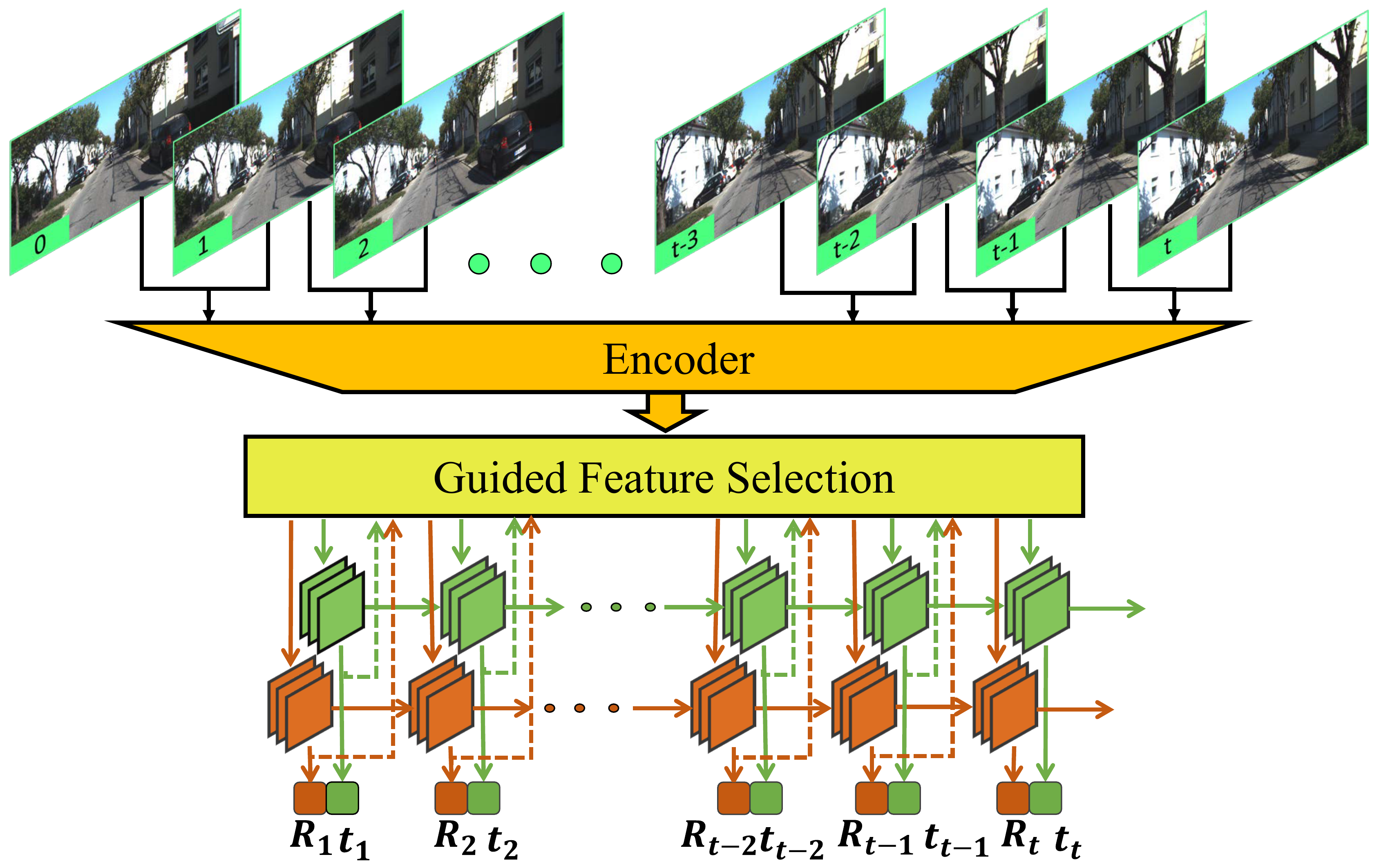}
	\caption{An overview of our architecture. Rotation and translation are estimated separately in a dual-branch recurrent network. The specific motions are calculated using corresponding features selected with the guidance of previous output.}
	%  through a context-aware feature selection module
	\label{fig:structure}
\end{figure}

\begin{itemize}
	\item We propose a dual-branch recurrent network with convolutional structure underneath for decoupled camera pose estimation, enabling the model to learn different motion patterns via specific features. 
	%To the best of our knowledge, we are the first to realize this based on deep learning.
	
	\item We incorporate a context-aware feature selection mechanism to steer the network explicitly for distilling motion-sensitive information for each branch, using previous output as guidance spatially and temporally.
	
	\item Our experiments on the public benchmarks show that the proposed approach outperforms state-of-the-art VO methods for both joint and decoupled camera pose prediction.
	
\end{itemize}

The rest of this paper is organized as follows. In \S~\ref{related_work}, related works on monocular VO and context-aware learning strategy are discussed. In \S~\ref{approach}, we introduce the architecture of our Guided Feature Selection for Deep Visual Odometry. The performance of the proposed approach is compared with other state-of-the-art methods in \S~\ref{experiments}. We conclude the paper in \S~\ref{conclusion}.

%------------------------------------------------------------------------- 
\section{Related Work}
\label{related_work}
\subsection{Visual Odometry Based on Joint Pose Estimation}
Traditionally, VO algorithms can be roughly categorized into feature-based and direct methods. Feature-based approaches establish correspondences across images via  keypoints. VISO2 \cite{geiger2011stereoscan} utilizes circle matching between consecutive frames to realize an efficient monocular VO system. Since outliers and noises are unavoidable, all VO algorithms suffer from scale-drift and error accumulation. The problems can be partially solved in SLAM algorithms such as ORB-SLAM \cite{mur2017orb-slam2} by introducing pose graph optimization. Feature-based methods suffer from heavy time cost for feature extraction, and can fail in environments with limited texture information. Direct methods \cite{engel2014lsd-slam, engel2017dso} recover poses by directly minimizing photometric error. These methods do not require expensive feature extraction, yet are sensitive to illumination variations. DSO \cite{engel2017dso} alleviates this problem by integrating a full photometric calibration. Up to now, both feature-based and direct methods are designed for static scenes and may face problems encountering dynamic objects. Moreover, absolute scale cannot be recovered in these methods without auxiliary information.

Recently, due to the advances of deep learning for computer vision tasks, CNNs and RNNs have been utilized for pose estimation. DeMoN \cite{ummenhofer2017demon} estimates depth and motion from two consecutive images captured by monocular cameras. SfmLearner \cite{zhou2017egomotion} and its successors \cite{yin2018geonet, li2017undeepvo} recover depth of scenes and ego-motions from unlabeled sequences with view synthesis as supervisory signal. DeepVO \cite{wang2017deepvo} learns camera poses from image sequences by combining CNNs and RNNs. It feeds 1D vectors learned by an encoder into a two-layer regular LSTM to predict motion of each frame and builds the loss function over the absolute joint poses at each time step. ESP-VO \cite{wang2017espvo} extends DeepVO by inferring poses and uncertainties directly in a unified framework. VINet \cite{clark2017vinet} fuses visual and inertial information in an intermediate representation level to eliminate manual synchronizations and performs sequence-to-sequence learning. 

The methods above, however, consider measly the response of visual cues to different motion types. Besides, spatial connection is ignored in approaches based on regular RNNs, such as DeepVO and ESP-VO.

%-------------------------------------------------------------------------
\subsection{Visual Odometry Based on Decoupled Pose Estimation}

Generally, instead of sharing the same features with translation, rotation can be recovered via geometric priors of certain scenes. Vanishing point \cite{lee2015real, jo2018camera} and planar structure \cite{kim2017visual, straub2015real, dvo2013robust, mwo2016zhou} are two kinds of frequently-used visual cues. \cite{straub2015real, dvo2013robust, mwo2016zhou} decouple the rotation and translation to estimate orientation by tracking Manhattan frames. \cite{kim2017visual} extends to compute translational motion in VO system by minimizing de-rotated reprojection error given the rotation. \cite{bazin2010motion} exploits vanishing points to recover the absolute attitude, and uses a 2-point algorithm to estimate translation for catadioptric vision. \cite{kaess2009flow, tardif2008monocular} select features for specific motion estimation according to depth values, since points at infinity are hardly influenced by translation,  and hence are appropriate to estimate orientation. The strategy is also adopted in stereo SLAM systems \cite{mur2017orb-slam2}.

Methods relying on Manhattan World assumption or depth of features achieve promising results in limited scenes but at a cost of reduced generalization and heavy noise. Instead, our method partially solves these problems by leveraging CNNs to extract features explicitly, and effectively.   
% While the aforementioned approaches break the reliance of Manhattan world assumption, it leads large error due to the noise in disparity computation. 

\subsection{Context-Aware Learning Mechanism}
Contextual information is helpful in improving the performance of networks. It has been widely utilized in many computer vision tasks. Specifically, TRACA \cite{choi2018context} uses the context of coarse category of tracking targets and proposes multiple expert auto-encoders to construct context-aware correlation filter for real-time tracking. PiCANet \cite{liu2017picanet} learns to selectively attend informative context locations for each pixel to generate contextual attention maps. EncNet \cite{zhang2018context} uses the semantic context to selectively highlight the class-dependent feature-maps for semantic segmentation. CEN \cite{mac2018context} defines the context as attributes assigned to each image and model the bias for image embeddings. 

Our model benefits from the \textit{small motion} between two consecutive views in an image sequence and exploits context i.e. continuity of neighboring frames in content and motion, to infer camera poses in a guided manner.

% \clearpage
%-------------------------------------------------------------------------
\section{Guided Feature Selection for Deep Visual Odometry}
\label{approach}

In this section, we introduce our framework (Fig.~\ref{fig:structure}) in detail. First, the model encodes RGB images to high-level features in \S~\ref{extraction}. Then, a context-aware motion guidance is adopted to recalibrate these features in \S~\ref{separation}. After that, the feature-maps are fed into two branches for learning rotation and translation in \S~\ref{guidance}. Finally, we design a loss function considering both the rotational and translational errors in \S~\ref{loss}.

%-------------------------------------------------------------------------
\subsection{Feature Extraction}
\label{extraction}
We harness the CNN to learn feature representation. Recently, plenty of excellent deep neural networks have been developed to deal with computer vision tasks such as classification \cite{hu2018squeeze}, objection detection \cite{he2017mask_rnn}, semantic segmentation \cite{chen2017deeplab} by focusing on appearance and content of images. The VO task, however, depends on geometrical information in input sequences. By taking the efficiency of transfer learning \cite{zamir2018taskonomy} into consideration, we build the encoder based on Flownet \cite{dosovitskiy2015flownet} proposed for optical flow estimation. We retain the first 9 convolutional layers, as \cite{wang2017deepvo, wang2017espvo}, encoding a pair of images into a 1024-channel stacked 2D feature-maps. The process can be described as
\begin{align}
\mathbf{X}_t = \mathcal{F}(I_{t-1}, I_{t}; \theta_\mathcal{F}) \ , \label{eq:encoder} 
\end{align}

\noindent where $I_{t-1}, I_{t}$ are consecutive frames. $\mathbf{X}_t=[\mathbf{X}_t^1, \mathbf{X}_t^2, ..., \mathbf{X}_t^C] \in \mathbb{R}^{H \times W \times C}$ is the extracted features with channel $C$ and size $H \times W$. $\mathcal{F}$ maps raw images to high-level abstract 3D tensors through parameters $\theta_\mathcal{F}$. Different from DeepVO \cite{wang2017deepvo} and ESP-VO \cite{wang2017espvo}, we keep the structure of feature-maps for retaining the spatial formulation rather than compressing features into 1D vectors. 

%-------------------------------------------------------------------------
\begin{figure}[t]
	\centering
	\subfigure[Vanilla model]{
		\begin{minipage}{0.45\textwidth}
			\centering
			\includegraphics[height=5.2cm]{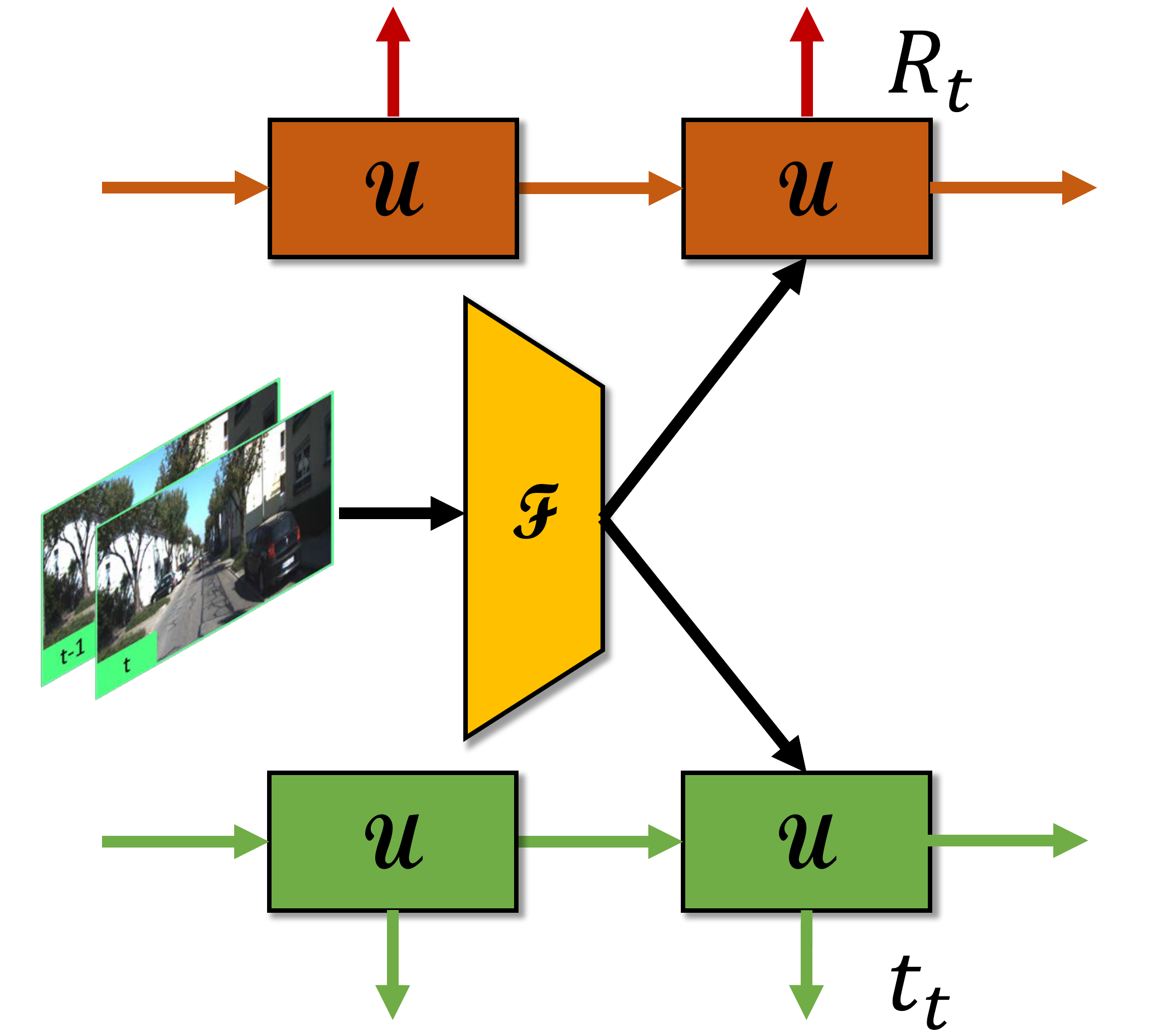}
			\label{fig:vanilla}
	\end{minipage}}
	%\quad
	\subfigure[Guidance model]{
		\begin{minipage}{0.45\textwidth}
			\centering
			\includegraphics[height=5.2cm]{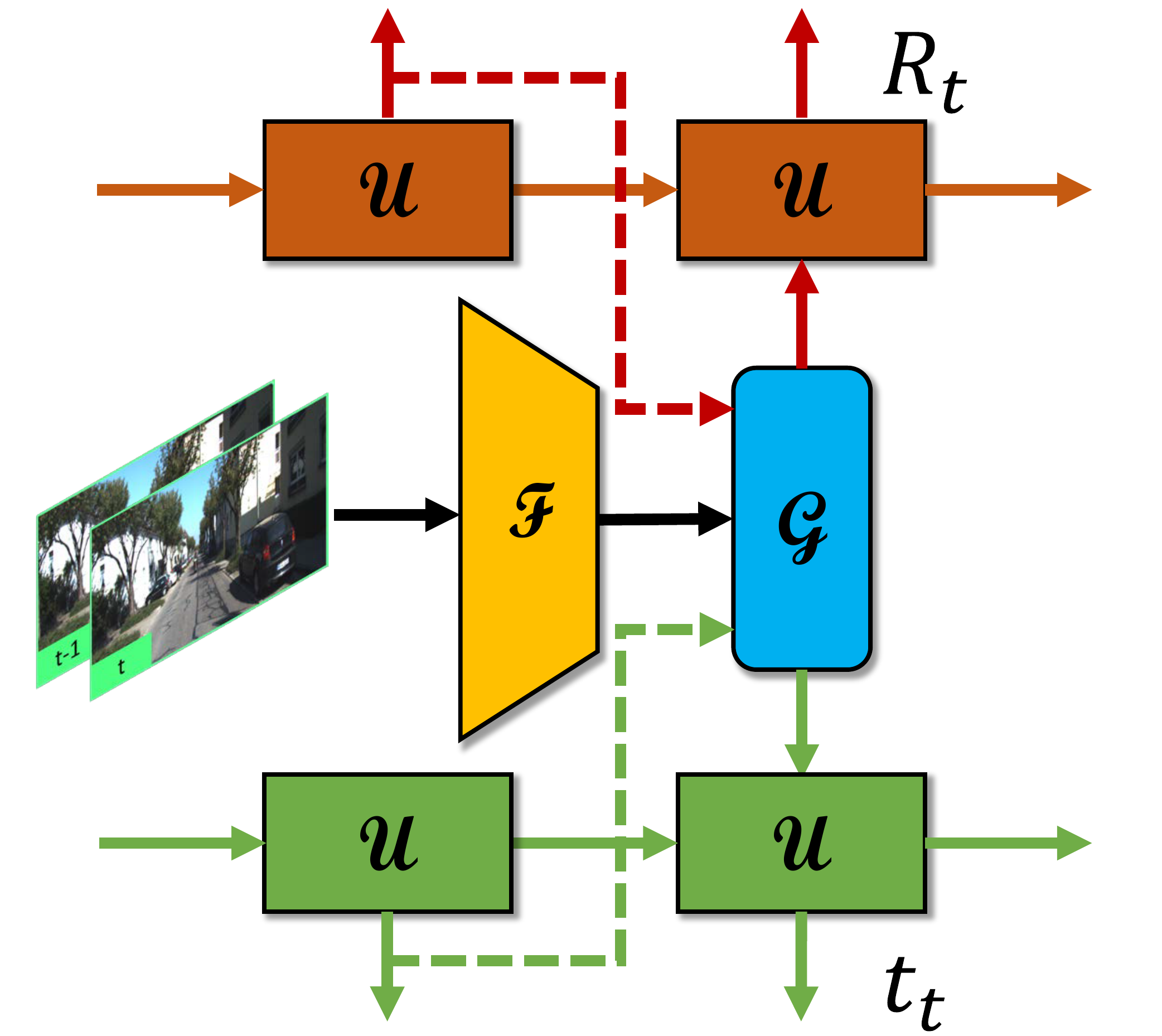}
			\label{fig:guidance} 
			%\hspace{0.3cm}
	\end{minipage}}
	\caption{An illustration of two structures for decoupled motion estimation. A vanilla structure (a) feeds feature-maps into a ConvLSTM unit directly, while the guided model (b) utilizes previous output as guidance for feature selection.}
	\label{fig:gru_fusion}
\end{figure}

\subsection{Dual-Branch Recurrent Network for Motion Separation}
\label{separation}

VO algorithms aim to recover camera poses from image sequences by leveraging the overlap between two or several consecutive frames. It's reasonable to model the sequences via LSTM \cite{hochreiter1997lstm}, a variation of RNN. In this case, the feature flow passing through recurrent units, carries rich accumulated information from previous observations to infer current output. Unfortunately, standard units of LSTM utilized by DeepVO \cite{wang2017deepvo} and ESP-VO \cite{wang2017espvo} requires 1D vector as input, and thus break the spatial structure of features. We rather adopt ConvLSTM \cite{xingjian2015convolutional}, an extended LSTM unit with convolution embedded preserving more detailed visual cues to form a two-branch recurrent model. Since \textit{gates} in ConvLSTM unit such as \textit{output gate, input gate, forget gate} can be thought intuitively as regulators of the flow of values going through the connections, features are filtrated and reorganized to fit relevant motions. The process can be controlled by
\begin{align}
\mathbf{O}_t, \mathbf{H}_t = \mathcal{U}(\mathbf{X}_t, \mathbf{H}_{t-1}; \theta_\mathcal{U}) \ , \label{eq:rnn}
\end{align}

\noindent where $\mathbf{X}_t, \mathbf{O}_t$ and $\mathbf{H}_t$ denote the input, output and hidden state at current time point, respectively. $\mathbf{H}_{t-1}$ is the previous hidden state. Note that $\mathbf{X}_t, \mathbf{O}_t$ are both 3D tensors, so are the hidden states $\mathbf{H}_t, \mathbf{H}_{t-1}$. $\mathcal{U}$ plays the role of recurrent units of ConvLSTM \cite{xingjian2015convolutional} with $\theta_\mathcal{U}$ representing its parameters. 

We create a two-branch recurrent model for decoupled motion prediction, enabling each branch to control corresponding feature flow for one type of motion. In general, a vanilla model feeds extracted features $\mathbf{X}_t$ into each branch directly, using the recurrent units for information selection, as show in Fig.~\ref{fig:vanilla}. The unit, however, may be inefficient in feature selection due to finite capacity. Amount of redundant information may ulteriorly aggravates the situation, and hence degrades the accuracy. 

Intuitively, previous output contains valuable visual cues for corresponding motion estimation, and thus can serve as a supervisor. In our model, raw feature-maps $\mathbf{X}_t$ are reconsidered before fed into each branch as depicted in Fig.~\ref{fig:guidance}. Thereby, each motion pattern can be learned from most related features, as discussed in \S~\ref{guidance}. The process can be described as 
\begin{align}
\mathbf{O}^{motion}_t, \mathbf{H}^{motion}_t = \mathcal{U}(\mathbf{X}^{motion}_t, \mathbf{H}^{motion}_{t-1}; \theta_\mathcal{U}) \ , \label{eq:seq_rnn}
\end{align}

\noindent where $\mathbf{X}_t^{motion}, \mathbf{H}_t^{motion}$ and $\mathbf{O}_t^{motion}$ denote the recalibrated features, hidden state and output for specific \textit{motion}. $\mathbf{H}_{t-1}^{motion}$ is the previous hidden state for the $motion$ branch. Here, \textit{motion} indicates \textit{rotation} and \textit{translation} specifically in our work. Obviously, the model is flexible to accept various motion patterns according to tasks.

%-------------------------------------------------------------------------
\begin{figure}[t]
	\centering
	\includegraphics[height=4.5cm]{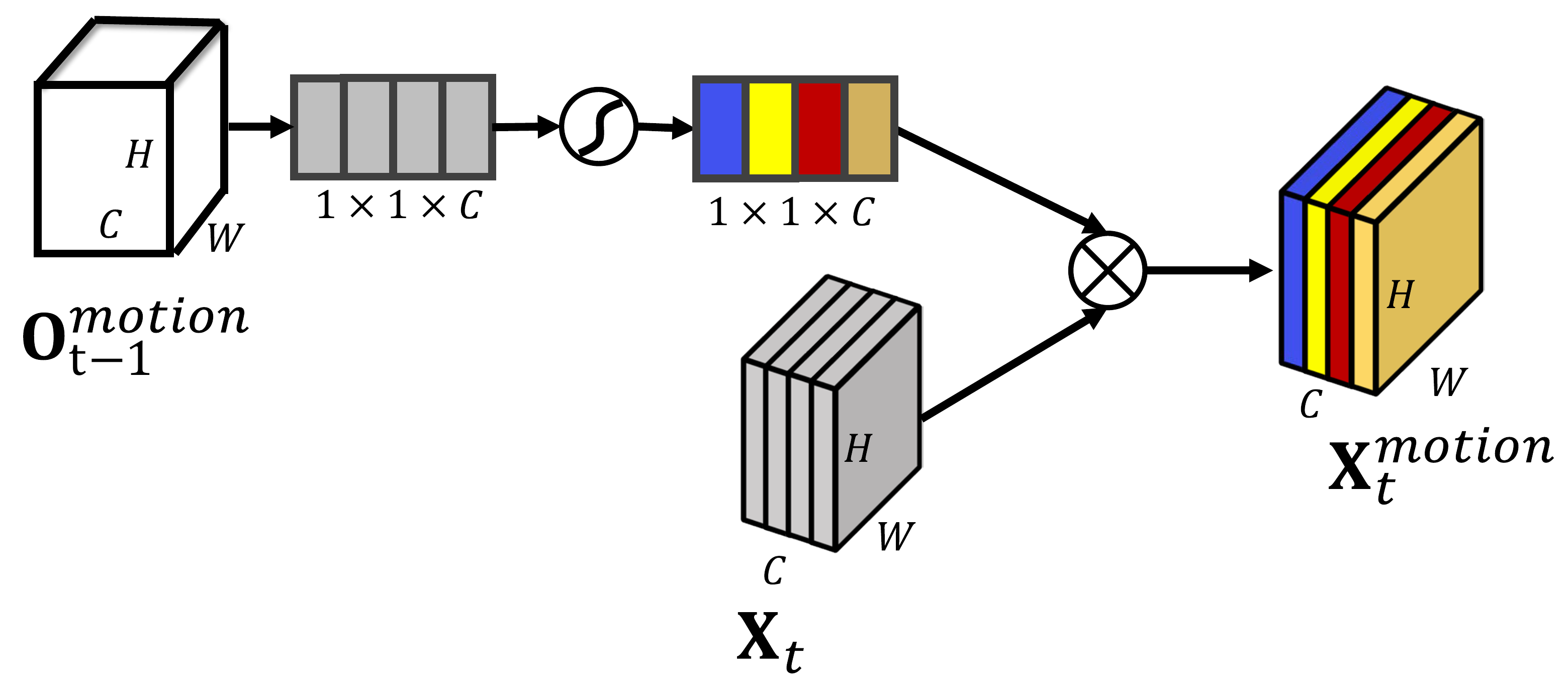}
	\caption{SENet-like guidance module. Encoded features are scaled along channel dimension according to only previous output without the participation of current input.}
	\label{fig:guidance_senet}
\end{figure}

\subsection{Guided Feature Selection}
\label{guidance}
To achieve the purpose of generating related information for each branch, our approach benefits from \textit{small} motion between neighboring views by incorporating a guidance module to selectively distill features for current pose inference adaptively as 
\begin{align}
\mathbf{X}_t^{motion} & = \mathcal{G}(\mathbf{X}_t, \mathbf{O}_{t-1}^{motion}; \theta_\mathcal{G}) \ .
\label{eq:sep}
\end{align}

Here, $\mathcal{G}$ is a function that maps features $\mathbf{X}_t$ to motion-sensitive tensors $\mathbf{X}_t^{motion}$ with the supervision of previous output $\mathbf{O}_{t-1}^{motion}$. $\theta_\mathcal{G}$ denotes the weights of $\mathcal{G}$. We introduce two strategies for context-aware guidance considering the connections in temporal and spatial domain. The first one is a SENet-like guidance, and the second one is a correlation-based guidance.  

\textbf{SENet-like guidance.} Inspired by the work of SENet \cite{hu2018squeeze}, in which a \textit{Squeeze-and-Extinction} block is implemented to \textit{self-recalibrate} channel-wise feature responses. Rather than self-adjusting the weights, we focus on the relationship in temporal domain. The previous output is first passed through a global average pooling (GAP) layer to yield a channel-wise descriptor. Two Fully-Connected (FC) layers are followed to learn inner channel dependence and produce scale values. Then, a sigmoid layer normalizes these values to [0, 1]. The final output is obtained by rescaling features along the channel dimension. A diagram of channel-wise guidance is shown in Fig.~\ref{fig:guidance_senet}. The process is formulated as
\begin{align}
\mathbf{s}^{motion} &= \sigma (\mathbf{W}_2\delta(\mathbf{W}_1GAP(\mathbf{O}_{t-1}^{motion}) + \mathbf{b}_1) + \mathbf{b}_2) \ , \label{eq:scale} \\
\mathbf{X}_t^{motion,c} &= \mathbf{X}_t^c \cdot s_c^{motion} \ , \label{eq:mul_scale}
\end{align}

\noindent where $\mathbf{W}_1, \mathbf{W}_2$ denote the two FC layers with $\mathbf{b}_1, \mathbf{b}_2$ as biases. $\sigma, \delta$ indicate the sigmoid and ReLU \cite{relu2010} activation functions. $\mathbf{s}^{motion} = [s_1^{motion}, s_2^{motion}, ..., s_C^{motion}]$ is the obtained scale vectors. $\mathbf{X}_t^c, \mathbf{X}_t^{motion, c}$ are feature-maps of $\mathbf{X}_t$ and $\mathbf{X}_t^{motion}$ of channel $c$.

Note that, SENet aims to exploit contextual interdependencies of $\mathbf{X}_t$, while our algorithm focuses on temporal consistency by filtering $\mathbf{X}_t$ according to the proposal of $\mathbf{O}_{t-1}^{motion}$. 

\begin{figure}[t]
	\centering
	\subfigure[Point-wise correlation]{
		\begin{minipage}{0.48\textwidth}
			\centering
			\includegraphics[height=2.8cm]{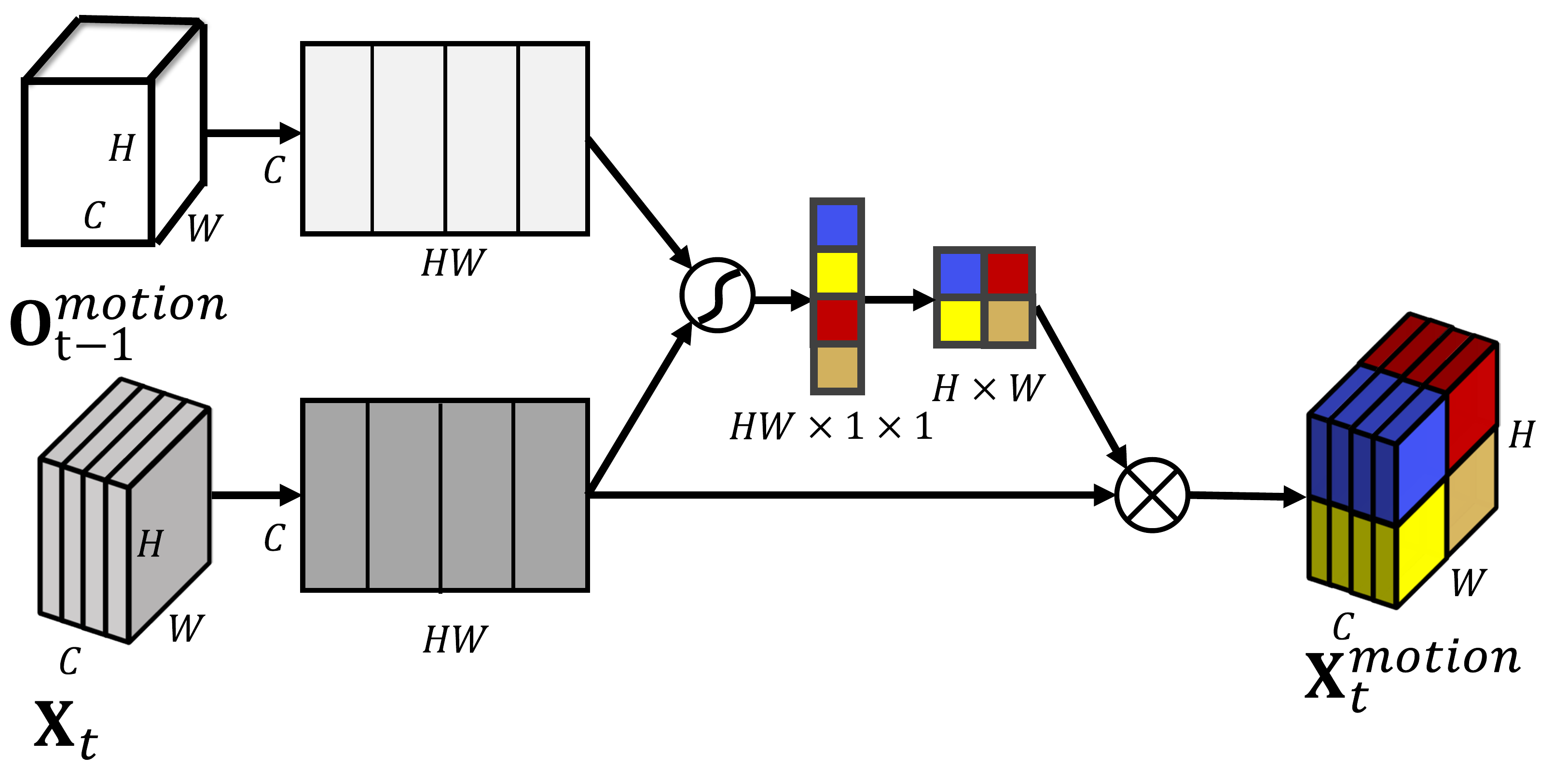}
			\label{fig:guidance_channel}
	\end{minipage}}
	\subfigure[Channel-wise correlation]{
		\begin{minipage}{0.48\textwidth}
			\centering
			\includegraphics[height=2.8cm]{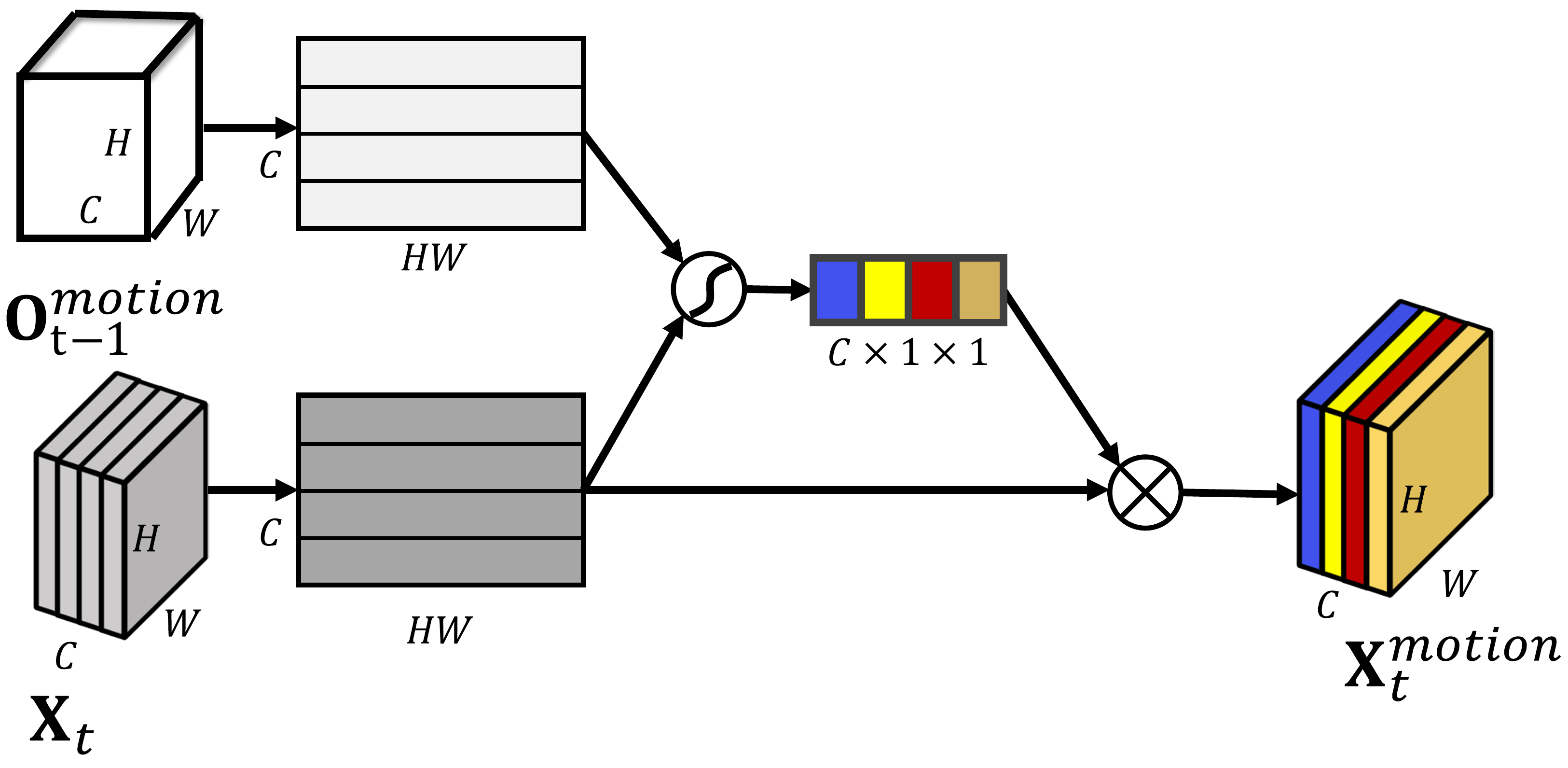}
			\label{fig:guidance_pixel}
	\end{minipage}}
	\caption{An illustration of two types of correlation-based guidance. Point-wise correlation (a) takes values at the same position of all 2D feature-maps as a unity, while channel-wise correlation (b) computes the weight for feature-map of each channel.}
	\label{fig:guidance_context}
\end{figure}

\textbf{Correlation-based guidance.} SENet-like subnetwork produces relatively coarse scalars in temporal domain without considering spatial relationship between $\mathbf{O}_{t-1}$ and  $\mathbf{X}_t$. Since the detail information is not kept, the performance is not satisfying. We further explore the guidance at a finer level from the aspect of correlation between $\mathbf{O}_{t-1}^{motion}$ and $\mathbf{X}_t$, as depicted in Fig.~\ref{fig:guidance_context}.

Since $\mathbf{X}_t$ and $\mathbf{O}_{t-1}^{motion}$ are both 3D tensors of stacked 2D-features, there are two different approaches to calculating the cross-correlation parameters by taking each pixel position along channel dimension as a column (Fig.~\ref{fig:guidance_channel}) or each feature-map as a unity (Fig.~\ref{fig:guidance_pixel}). In the first form, we compute the cosine similarity of each corresponding column first, and normalize the weight next. We have tried the sigmoid and softmax for normalization. The sigmoid function gives better performance in our experiments. The process can be described as
\begin{align}
s_{(u,v)}^{motion} &= \sigma(\frac{\mathbf{X}_{t, (u,v)} \cdot \mathbf{O}_{t-1, (u,v)}^{motion}}{||\mathbf{X}_{t, (u,v)}||_2 \cdot ||\mathbf{O}_{t-1, (u,v)}^{motion}||_2}) \ , \label{eq:pixel} \\
\mathbf{X}_{t, (u,v)}^{motion} &= \mathbf{X}_{t, (u,v)} \cdot s_{(u,v)}^{motion} \ ,  \label{eq:scale_pixel}
\end{align}

\noindent here $\mathbf{X}_{t, (u,v)}$ and $\mathbf{O}_{t-1, (u,v)}^{motion}$ are vectors with size of $C$ at each point of $\mathbf{X}_t$ and $\mathbf{O}_{t-1}^{motion}$ indexed by $u, v$. $s_{(u,v)}^{motion}$ is the scale for re-weighted tensor $\mathbf{X}_{t}^{motion}$ at $(u, v)$. Intuitively, if current vectorial feature $\mathbf{X}_{t, (u,v)}$ is close to previous output $\mathbf{O}_{t-1, (u,v)}^{motion}$, it should be assigned a larger weight, otherwise a smaller one.  

In the second type, 2D feature map of each channel is unified as a vector, on which we compute the correlation as
\begin{align}
s_c^{motion} = \sigma(\frac{Vec(\mathbf{X}_t^c) \cdot Vec(\mathbf{O}_{t-1}^{motion,c})}{||Vec(\mathbf{X}_t^c)||_2 \cdot || Vec(\mathbf{O}_{t-1}^{motion, c})||_2}) \ ,\label{eq:channel}
\end{align}

\noindent where $Vec(\cdot)$  reshapes a 2D feature map into a vector for correlation computation. $\mathbf{X}_t$ is re-weighted adaptively for each branch according to the correlation parameters as (\ref{eq:mul_scale}).

Context-aware motion guidance scheme brings better performance for our model with a limited time cost. We analyze the boosted efficiency in \S~\ref{experiments}.  
% possess

%-------------------------------------------------------------------------
\subsection{Loss Function}   
\label{loss}
Our architecture learns rotation and translation in two individual recurrent branches separately, hence the final loss consists of both rotational and translational errors. We define the loss on the absolute pose error of each view using the $L_2$ norm. The loss functions are formulated as 
\begin{align}
\mathcal{L}_i^{rot} & =  ||\hat{\bm{\phi}}_{i} - \bm{\phi}_{i}||_2 \ , \label{eq:rot} \\
\mathcal{L}_i^{trans} & = ||\hat{\bm{p}}_{i} - \bm{p}_{i}||_2 \ , \label{eq:trans} \\ 
\mathcal{L}_{total} & = \sum_{i=1}^{t} \frac{1}{i}(\mathcal{L}_i^{trans} + k\mathcal{L}_i^{rot}) \ . \label{eq:total}
\end{align} 

Here $\hat{\bm{p}}_{i}, \bm{p}_{i}, \hat{\bm{\phi}}_{i} $, and $\bm{\phi}_{i}$ represent the predicted and ground-truth translation and rotation (Euler angles) of the $i$-th view in world coordinate. $\mathcal{L}_i^{rot}$ and $\mathcal{L}_i^{trans}$ denote the rotational and translational error of the $i$-th frame, respectively. The final loss, $\mathcal{L}_{total}$ sums the averaged loss of each time step. $t$ is the current frame index in a sequence. $k$ is a fixed parameter for balancing the rotational and translational errors. It is set to 100 and 10 in experiments on KITTI and ICL\_NUIM dataset respectively.

\section{Experiments}
\label{experiments}
We first discuss the implementation details of our network in \S~\ref{implementation}, and introduce the datasets used in \S~\ref{dataset}. We compare the effectiveness of variations of our network, RNN for the regular recurrent network, SRNN for the dual-branch recurrent model, SRNN\_se for dual-branch network plus senet-like contextual mechanism, SRNN\_channel for dual-branch network plus channel-wise correlation, and SRNN\_point for dual-branch network plus point-wise correlation, in \S~\ref{result_attention}. Next, we compare our proposal with current methods on the KITTI dataset \cite{geiger2012kitti} in \S~\ref{result_kitti_dataset}, and ICL\_NUIM dataset \cite{icl2014benchmark} in \S~\ref{result_icl_dataset}.

\subsection{Implementation}
\label{implementation}
\textbf{Training.} Our model takes monocular image sequences as input. The image size can be arbitrary because the model has no requirement of compressing images into vectors.  We use 7 consecutive frames to construct a sequence considering the time cost, yet our model can accept dynamic lengths of inputs. \\
%The rotation and translation branches estimate Euler angles and translations, both of which are presented as 3DoF parameters, respectively. \newline
\textbf{Network.} Weights of recurrent units are initialized with MSRA \cite{he2015msra}, while the encoder is based on pre-trained Flownet \cite{dosovitskiy2015flownet} to speed up convergence. Our networks are implemented by PyTorch \cite{pytorch} on an NVIDIA 1080Ti GPU. We employ the poly learning rate policy \cite{chen2017deeplab} with power = $0.9$ and initial learning rate = $10^{-4}$. Adam \cite{Kingma2014Adam} with $\beta_1=0.9, \beta_2=0.99$ is used as optimizer. The networks are trained with a batch size of 4, a weight decay of $10^{-4}$ for 150,000 iterations in total. 

\input{fig_table/fig_error_each_view}

\subsection{Dataset}
\label{dataset}
\textbf{KITTI.} The public KITTI dataset is used by both model- \cite{mur2017orb-slam2, geiger2011stereoscan} and learning-based methods \cite{zhou2017egomotion, wang2017deepvo, wang2017espvo}. The dataset consists of 22 sequences captured in urban and highway environments at a relatively low sample frequency (10 fps) at speed up to 90km/h. Seq 00-10 provide raw data with ground-truth represented as 6DoF motion parameters by considering the complicate urban environments,  while Seq 11-21 provide only raw sensor data. In our experiments, the left RGB images are resized to 1280 x 384 for training and testing. We adopt the same train/test split as DeepVO \cite{wang2017deepvo} by using Seq 00, 02, 08, 09 for training and Seq 03, 04, 05, 06, 07 and 10 for quantitative evaluation.  \newline
%Many dynamic objects such as cars, pedestrians make the dataset very challenging for monocular VO algorithms.
\textbf{ICL\_NUIM.} The ICL\_NUIM dataset \cite{icl2014benchmark} consists of 8 sequences of RGB-D images captured within synthetically generated living room and office. Images in this dataset meet the Manhattan World assumption. The dataset is widely used for VO/SLAM \cite{kim2017visual, mwo2016zhou, dvo2013robust} and 3D reconstruction \cite{dai2017bundlefusion}. ICL\_NUIM dataset is synthesized by a full 6DoF handheld camera and thus is challenging for monocular VO methods due to complicated motion patterns. Our model is trained on kt0, kt3 and evaluated on kt1, kt2 on the living room and office datasets, respectively. Only RGB images with size of 640 x 480 are used in our experiments.

\input{fig_table/fig_attention}

\subsection{Evaluation of Context-Aware Mechanism}
\label{result_attention}
We first evaluate the efficiency of \textit{context-aware} strategies by analyzing rotational (Fig.~\ref{fig:variations_rotation}) and translational (Fig.~\ref{fig:variations_translation}) errors along each view of the sequence on KITTI test datasets. We adopt the orientation and position drift errors divided by traveling length as metric. In Fig.~\ref{fig:result_attention}, we observe that results of the vanilla network are remarkably improved by the context-aware guidance, and meanwhile, networks with different contextual modules behave diversely.

Among the models with guidance, SRNN\_{se} extends the self-recalibration of SENet \cite{hu2018squeeze} by introducing temporal relationship, leading to improvement in decoupled motion learning. Compared with SRNN\_{se}, the superior performance of SRNN\_{point} and SRNN\_{channel} suggests that correlation may be more effective in feature filtration for the VO task. The results of SRNN\_{channel} are slightly better than SRNN\_{point}. We explain that keeping the interdependence of feature-map in each channel may be a better manner for the guidance.

\input{fig_table/table_kitti_00_10}
\subsection{Results on KITTI Dataset}
\label{result_kitti_dataset}
We compare our framework with model- and learning-based monocular VO methods on the KITTI test sequences. The error metrics, i.e., averaged Root Mean Square Errors (RMSEs) of the translational and rotational errors, are adopted for all the subsequences of lengths ranging from 100, 200 to 800 meters. 

Most monocular VO methods cannot recover absolute scale, and their results require post alignment with ground-truth. Therefore, the open-source VO library VISO2 \cite{geiger2011stereoscan} estimating scale according to the height of camera is adopted as the baseline method. The results of both monocular (VISO2-M) and stereo (VISO2-S) versions are provided. Table~\ref{table:table_kitti} indicates that our models, even the vanilla version, outperform VISO2-M in terms of both rotation and translation estimation by a large margin. Note that the scale is learned during the end-to-end training without any post alignment or relying on priori knowledge such as the height of camera used by VISO2-M. VISO2-S gains superior performance due to the advantages of stereo image pairs in scale recovery and data association. Note that, our SRNN\_channel achieves very close performance to VISO2-S provided solely monocular images. Qualitative comparisons are shown in Fig.~\ref{fig:trajectory_kitti_00_10}. Our approach outperforms VISO2-M especially in handling complicated motions. 

Besides, we compare our approach against current learning-based supervised methods DeepVO \cite{wang2017deepvo} and ESP-VO \cite{wang2017espvo}, both of which are implemented on a single branch with standard LSTM for coupled motion estimation. Table~\ref{table:table_kitti} illustrates the efficiency of our vanilla model. The improvement of this version is slight. We assume that extracting motion-sensitive features from encoded feature-maps directly may limit the accuracy. Fortunately, the deficiency is compensated by the context-aware feature selection mechanism. Our models with guidance outperform DeepVO and ESP-VO consistently. Meanwhile, out method achieves superior performance than unsupervised monocular approach, SfmLearner \cite{zhou2017egomotion}, and yields competitive results than stereo methods such as UnDeepVO \cite{li2017undeepvo} and Depth-VO-Feat \cite{zhan2018feature}.

\input{fig_table/fig_kitti_viso}

\input{fig_table/table_icl_rpe}
We intensively test our model in various scenes with complicate motions on Seq 11-21. The trajectories of results are illustrated in Fig.\ref{fig:trajectory_kitti_11-21}. In this case, our network is trained on all the training sequences (Seq 00-10), providing more data to avoid overfitting and maximize the ability of generation. We use the accurate VISO2-S \cite{geiger2011stereoscan} as reference due to the lack of ground-truth. Our model also obtains outstanding results. The appearing performance of our method reveals that the model generalizes well in unknown scenarios.

\subsection{Results on ICL\_NUIM Dataset}
\label{result_icl_dataset}
We further compare our model with methods on the challenging ICL\_NUIM dataset \cite{icl2014benchmark}.
%a public benchmark collected in man-made environments fulfilling the Manhattan World assumption. 
We test our networks on the living room and office datasets individually. The baseline methods include algorithms for both joint (DEMO \cite{demo2014zhang}, DVO \cite{dvo2013robust}) and separate (DRFE \cite{kim2017visual}, MWO \cite{mwo2016zhou}) pose recovery. The error metric for qualitative analysis is the root mean square error (RMSE) of the relative pose error (RPE), used in \cite{kim2017visual}. Table~\ref{table:icl}, indicates that the best performing SRNN\_channel yields lower errors in relative pose recovery on three among the four sequences. Note that all the four baselines use depth information while only monocular RGB images are used in our networks. The remarkable results suggest the potential ability of our proposal in dealing with more complicated motion patterns generated by handheld cameras or moving robotics.
% Training data in experiments may limit the performance of our model. However,

\input{fig_table/fig_kitti_11_21}

\section{Conclusions}
\label{conclusion}
In this paper, we propose a novel dual-branch recurrent neural network for decoupled camera pose estimation. The architecture is able to estimate different motion patterns via specific features. To enhance the performance, we incorporate a context-aware feature selection mechanism in both spatial and temporary domain, allowing the network to suppress useless information adaptively. We evaluate our techniques on the prevalent KITTI and ICL\_NUIM datasets, and results demonstrate that our method outperforms current state-of-the-art learning- and model-based monocular VO approaches for both joint and separate motion estimation. In the future, we plan to visualize the features used by specific motions. We will also explore the relationship between visual cues and more specific motion patterns such as rotation in different directions.

%\setlength\parindent{0em}
%\textbf{Acknowledgments.} This work is supported by the National Natural Science Foundation of China (61632003, 61771026), and National Key Research and Development Program of China (2017YFB1002601).

% In the future, we plan to visualize the features used by specific motions. We will also explore the relationship between visual cues and more specific motion patterns such as rotation in different directions.
%\clearpage
\bibliographystyle{splncs04}
\bibliography{egbib}

\end{document}

%% file: fig_table/fig_error_each_view.tex
\begin{figure}[t]
	\centering
	\subfigure[Rotational error of each view]{
		\begin{minipage}{0.48\textwidth}
			\centering
			\includegraphics[height=4.2cm]{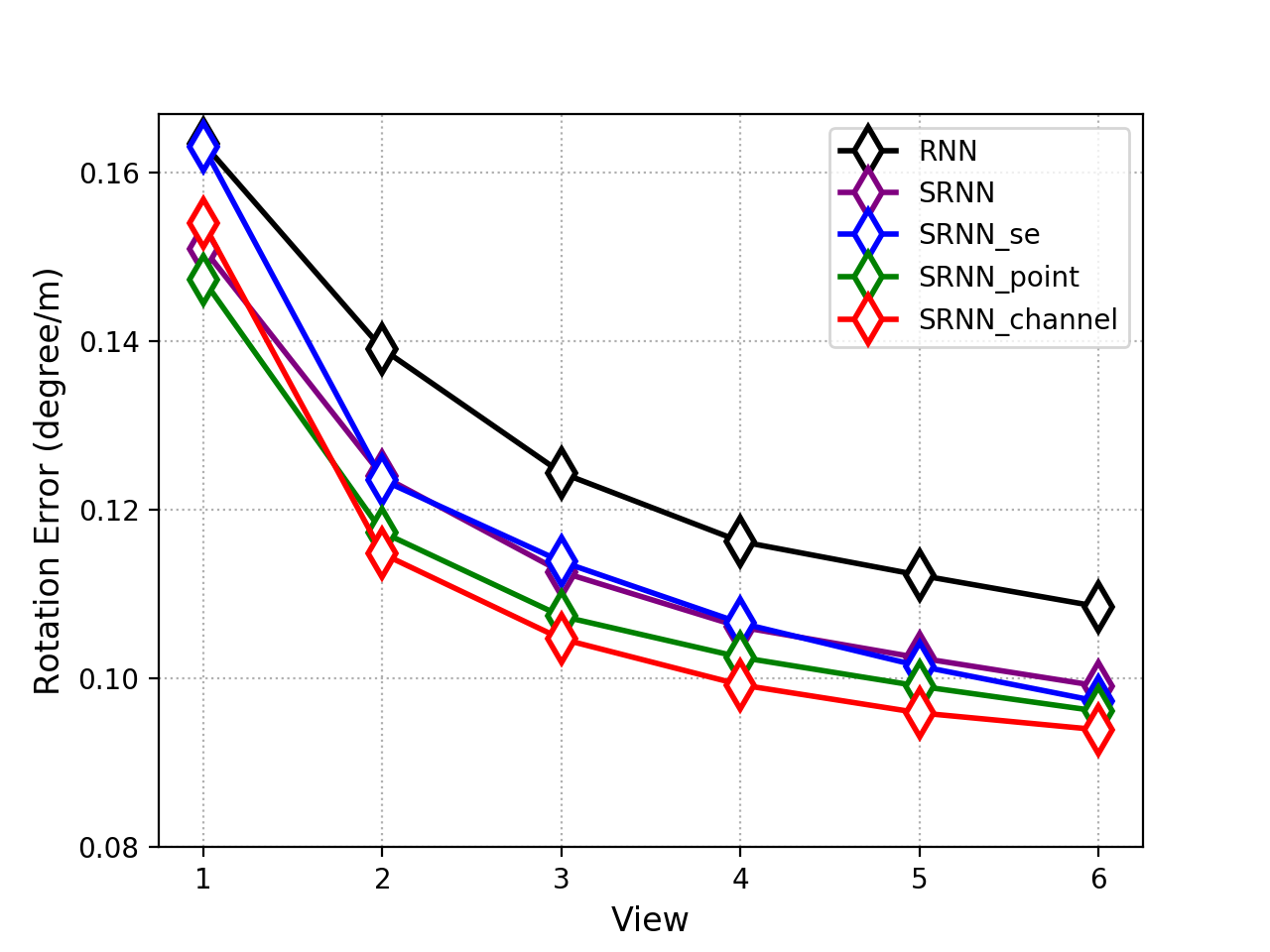}
			\label{fig:variations_rotation}
	\end{minipage}}
	\subfigure[Translational error of each view]{
		\begin{minipage}{0.48\textwidth}
			\centering
			\includegraphics[height=4.2cm]{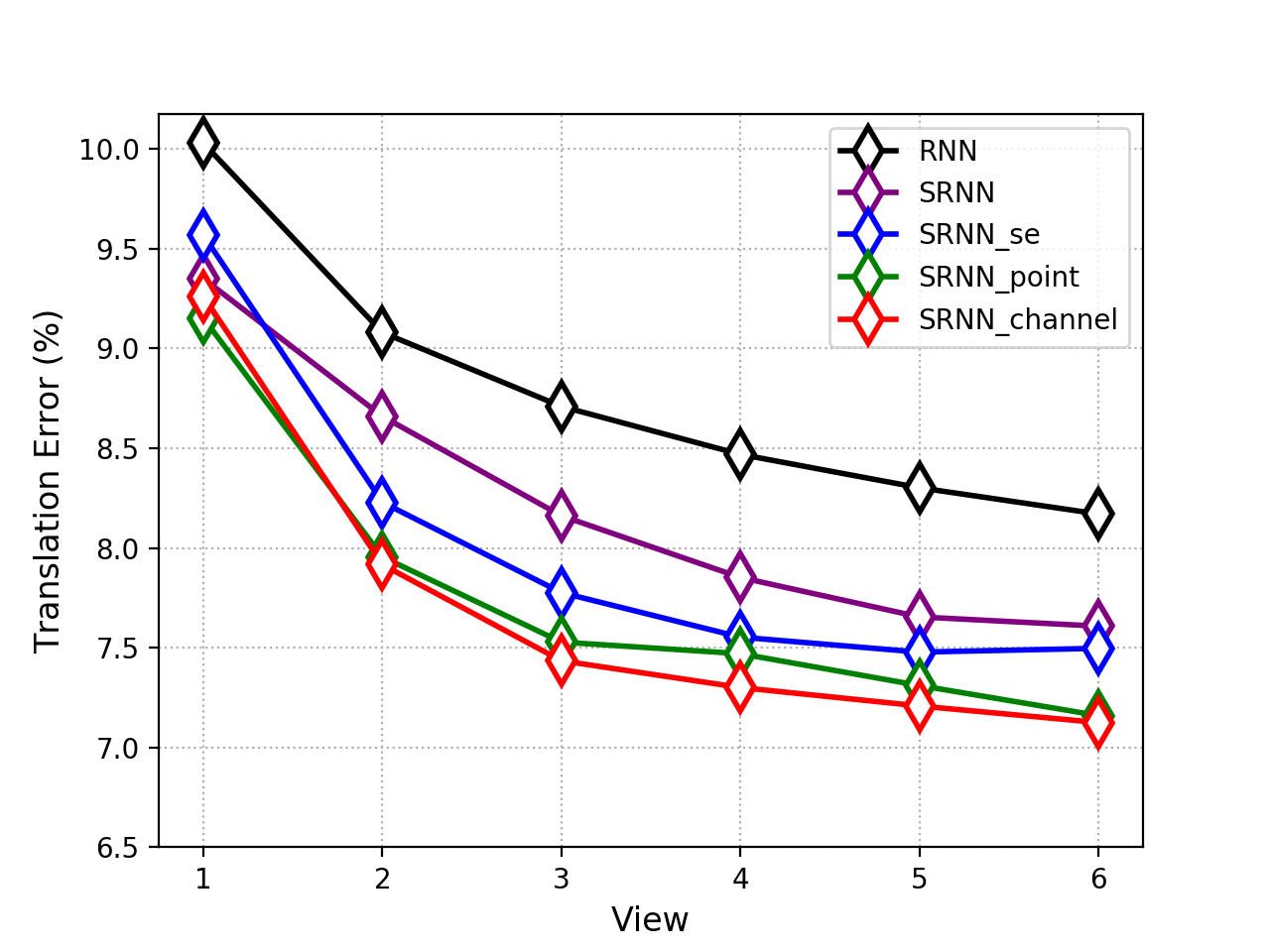}
			\label{fig:variations_translation}			
	\end{minipage}}
	\caption{Rotational (a) and translational (b) errors along each view of the network for joint motion prediction, vanilla and guided models for separate pose recovery.}
	\label{fig:result_attention}
\end{figure}
\setlength{\tabcolsep}{1.4pt}

%% file: fig_table/fig_attention.tex
\begin{figure}[t]
	\centering
	\subfigure[SRNN]{
		\begin{minipage}{0.48\textwidth}
			\centering
			\includegraphics[height=4.cm]{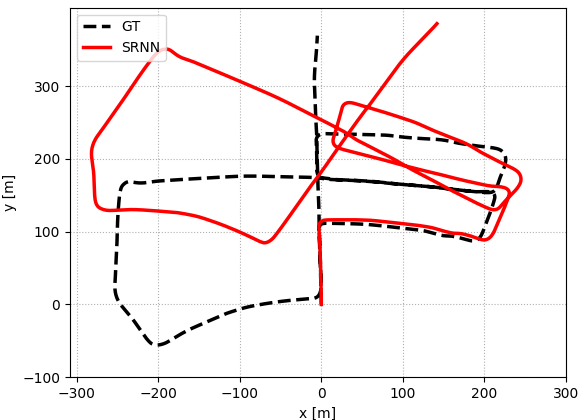}
			\label{fig:kitti10_a}
	\end{minipage}}
	\subfigure[SRNN\_se]{
		\begin{minipage}{0.48\textwidth}
			\centering
			\includegraphics[height=4.cm]{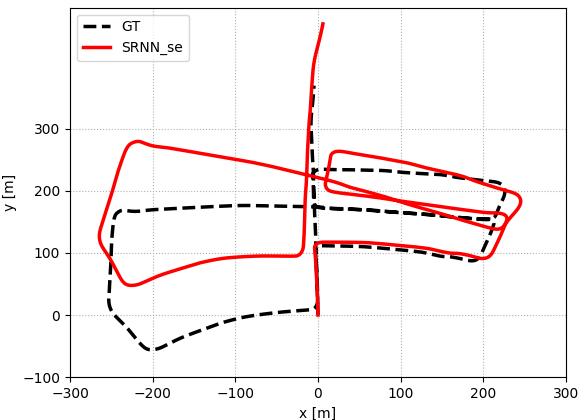}
			\label{fig:kitti10_b}
	\end{minipage}}
	\subfigure[SRNN\_point]{
		\begin{minipage}{0.48\textwidth}
			\centering
			\includegraphics[height=4.cm]{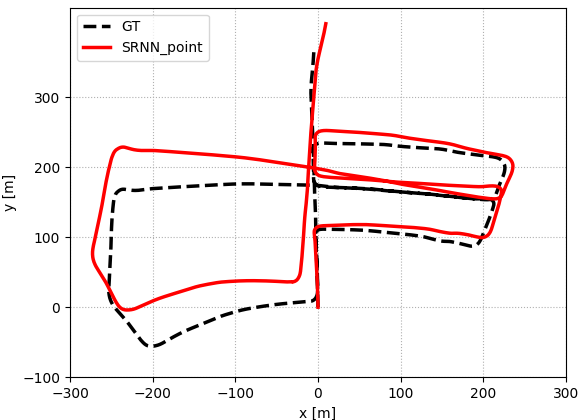}
			\label{fig:kitti10_c}
	\end{minipage}}
	%\qquad
	\subfigure[SRNN\_channel]{
		\begin{minipage}{0.48\textwidth}
			\centering
			\includegraphics[height=4.cm]{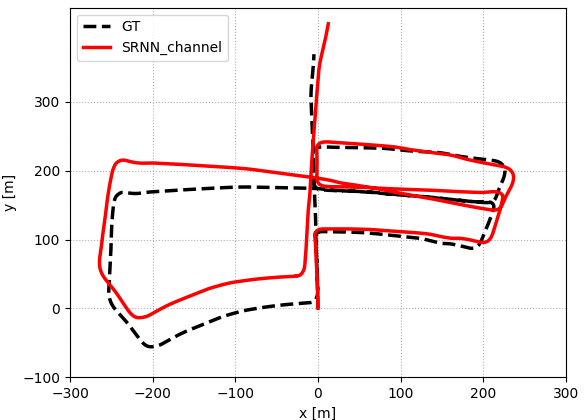}
			\label{fig:kitti10_d}
	\end{minipage}}
	\caption{Qualitative comparison of the trajectories of the vanilla and three guided models for separate motion estimation on the KITTI Seq 10.}
	\label{fig:variation_kitti10}
\end{figure}
\setlength{\tabcolsep}{1.4pt}

%% file: fig_table/table_kitti_00_10.tex
%We would be glad to respond to further questions and comments you may have.
%\tiny
\setlength{\tabcolsep}{2.4pt}
\begin{table}[t]
	\scriptsize
	\centering
		\caption{Results on the KITTI dataset. DeepVO \cite{wang2017deepvo}, ESP-VO \cite{wang2017espvo} and our models are trained on Seq 00, 02, 08 and 09. SfmLearner \cite{zhou2017egomotion}, UndeepVO \cite{li2017undeepvo} and Depth-VO-Feat \cite{zhan2018feature} are trained on Seq 00-08 in an unsupervised manner. The best results of monocular VO methods are highlighted without considering stereo ones including VISO2-S, UnDeepVO and Depth-VO-Feat. }
	%\begin{threeparttable}
		\begin{center}
			%\resizebox{\textwidth}{15mm}{
			\begin{tabular}{lclclclclclcl}
				\hline
				\hline
				& \multicolumn{12}{c}{Sequence} \\
				% \cline{2-15} 
				Method & \multicolumn{2}{c}{03} & \multicolumn{2}{c}{04} & \multicolumn{2}{c}{05} & \multicolumn{2}{c}{06} & \multicolumn{2}{c}{07} & \multicolumn{2}{c}{10} \\ 
				%Method & \multicolumn{2}{c}{$t_{rel} \quad r_{rel}$} &  \multicolumn{2}{c}{$t_{rel} \quad r_{rel}$} & \multicolumn{2}{c}{$t_{rel} \quad r_{rel}$} & \multicolumn{2}{c}{$t_{rel} \quad r_{rel}$} & \multicolumn{2}{c}{$t_{rel} \quad r_{rel}$} & \multicolumn{2}{c}{$t_{rel} \quad r_{rel}$} & \multicolumn{2}{c}{$t_{rel} \quad r_{rel}$} \\
				& $t_{rel}$ & $r_{rel}$ & $t_{rel}$ & $r_{rel}$ & $t_{rel}$ & $r_{rel}$ &  $t_{rel}$ & $r_{rel}$ &  $t_{rel}$ & $r_{rel}$ &  $t_{rel}$ & $r_{rel}$    \\
				\hline
				% DeepVO &  \multicolumn{2}{c} {$- \quad -$} &  \multicolumn{2}{c}{$8.49 \quad 6.89$} &  \multicolumn{2}{c}{$7.19 \quad 6.97$} &  \multicolumn{2}{c}{$2.62 \quad 3.61$} &  \multicolumn{2}{c}{$5.42 \quad 5.82$} &  \multicolumn{2}{c}{$3.91 \quad 4.60$} &  \multicolumn{2}{c}{$8.11 \quad 8.83$} \\
				%ORB\_SLAM-S [5] & 0.71 & 0.18 & 0.48 & 0.13 & 0.40 & 0.16 & 0.51 & 0.15 & 0.50 & 0.28 & 0.60 & 0.27 \\
				%ORB\_SLAM \cite{mur2017orb-slam2} & 0.93 & 0.20 & 0.88 & 0.21 & 12.05 & 0.22 & 19.59 & 0.23 & 13.63 & 0.37 & 24.97 & 0.30 \\
				%ORB\_SLAM-M [5] & 0.93 & 0.20 & 0.88 & 0.21 & 12.05 & 0.22 & 19.59 & 0.23 & 13.63 & 0.37 & 24.97 & 0.30 \\
				VISO2-S \cite{geiger2011stereoscan} & 3.21 & 3.25 & 2.12 & 2.12 & 1.53 & 1.60 & 1.48 & 1.58 & 1.85 & 1.91 & 1.17 & 1.30 \\
				UnDeepVO \cite{li2017undeepvo} & 5.00 & 6.17 & 5.49 & 2.13 & 3.40 & 1.50 & 6.20 & 1.98 & 3.15 & 2.48 & 10.63 & 4.65 \\
				Depth-VO-Feat \cite{zhan2018feature} & 15.58 & 10.69 & 2..92 & 2.06 & 4.94 & 2.35 & 5.80 & 2.07 & 6.48 & 3.60 & 12.45 & 3.46 \\
				\hline
				VISO2-M \cite{geiger2011stereoscan}   & 8.47 & 8.82 & 4.69 & 4.49 & 19.22 & 17.58 & 7.30 & 6.14 & 23.61 & 19.11 & 41.56 & 32.99 \\
				SfmLearner \cite{zhou2017egomotion} & 10.78 & 3.92 & 4.49 & 5.24 & 18.67 & 4.10 & 25.88 & 4.80 & 21.33 & 6.65 & 14.33 & 3.30 \\
				DeepVO \cite{wang2017deepvo} & 8.49 & 6.89 & 7.19 & 6.97 & $\mathbf{2.62}$ & 3.61 & $\mathbf{5.42}$ & 5.82 & 3.91 & 4.60 & 8.11 & 8.83 \\
				ESP-VO \cite{wang2017espvo} & 6.72 & 6.46 & 6.33 & 6.08 & 3.35 & 4.93 & 7.24 & 7.29 & 3.52 & 5.02 & 9.77 & 10.2 \\
				\hline
				%\textbf{RNN\_g} & 11.10 & 6.28 & 11.37 & 3.89 & 9.97 & 3.81 & 24.39 & 9.64 & 15.39 & 4.18 & 15.37 & 6.92 \\
				$\mathbf{RNN}$ & 6.36 & 3.62 & 5.95 & 2.36 & 5.85 & 2.55 & 14.58 & 4.98 & 5.88 & 2.64 & 7.44 & 3.19 \\
				%\textbf{RNN\_lg} & 7.72 & 4.43 &12.41 & $\mathbf{3.59}$ & 9.31 & 3.69 & 17.63 & 5.31 & 12.22 & 4.47 & 10.56 & $\mathbf{3.20}$ \\
				$\mathbf{SRNN}$ & 5.85 & 3.77 & 4.22 & 2.79 & 5.33 & 2.36 & 13.60 & 4.21 & 4.62 & 2.48 & 7.08 & 2.79 \\
				$\mathbf{SRNN\_se}$ & 5.45 & 3.33 & 4.11 & 1.70 & 4.74 & 2.21 & 12.44 & 4.45 & 4.23 & 2.67 & 6.79 & 2.91 \\
				%\textbf{BRNN\_g} &  $5.76$ & 3.71 & 12.44 & 4.34 & 8.88 & 3.59 & 22.54 & 8.55 & 11.97 & 4.19 & 11.06 & 4.45 \\
				$\mathbf{SRNN\_point}$ &  5.64 & $\mathbf{3.06}$ & 3.98 & 1.79 & 3.71 & 1.70 & 9.16 & 3.27 & 3.57 & 2.53 & 6.77 & 2.82 \\
				%\textbf{BRNN\_lg}  & $\mathbf{5.12}$ & $\mathbf{3.45}$ & 7.06 & $3.73$ & 7.53 & $\mathbf{3.24}$ & 14.49 & $\mathbf{5.22}$ & 5.28 & $\mathbf{2.30}$ & $\mathbf{7.24}$ & $3.36$ \\
				%\textbf{BRNN\_lg}  & $\mathbf{5.21}$ & $\mathbf{3.45}$ & 10.40 & 3.72 & 8.11 & $\mathbf{3.24}$ & 14.99 & $\mathbf{5.22}$ & 7.11 & $\mathbf{2.30}$ & 10.44 & 3.36 \\
				$\mathbf{SRNN\_channel}$  & $\mathbf{5.44}$ & 3.32 & $\mathbf{2.91}$ & $\mathbf{1.30}$ & 3.27 & $\mathbf{1.62}$ & 8.50 & $\mathbf{2.74}$ & $\mathbf{3.37}$ & $\mathbf{2.25}$ & $\mathbf{6.32}$ & $\mathbf{2.33}$ \\
				\hline
				\hline
			\end{tabular}
			\label{result_kitti}
			
			%\begin{left}
			\begin{tablenotes}
				%\left
				%\item $*$ORB\_SLAM \cite{mur2017orb-slam2} is not evaluated with other methods since it results are aligned with ground-truth using 7-DoF (rigid body + scale) transformation.
				\footnotesize
				\item $t_{rel}: $ average translational RMSE drift (\%) on length from 100, 200 to 800 m.
				\item $r_{rel}: $ average rotational RMSE drift (${}^{\circ}$/100m) on length from 100, 200 to 800 m.
				%\item Our models and DeepVO are trained on Sequence 00, 02, 08, and 09, while ESP-VO uses additional Sequence 01 for training.
			\end{tablenotes}
		\end{center}
		%\end{left}
	%\end{threeparttable}}
% The best results of monocular VO are highlighted.} %Predicted trajectories of ORB\_SLAM \cite{mur2017orb-slam2} are aligned with ground-truth due to unknown scale. VISO2\_M \cite{geiger2011stereoscan} recovers scale using the fixed camera height as priori.}
	\label{table:table_kitti}
\end{table}
\setlength{\tabcolsep}{1.4pt}

%% file: fig_table/fig_kitti_viso.tex
\begin{figure}[t]
	\centering
	\subfigure[Seq 03]{
		\begin{minipage}{0.48\textwidth}
			\centering
			\includegraphics[height=4.cm]{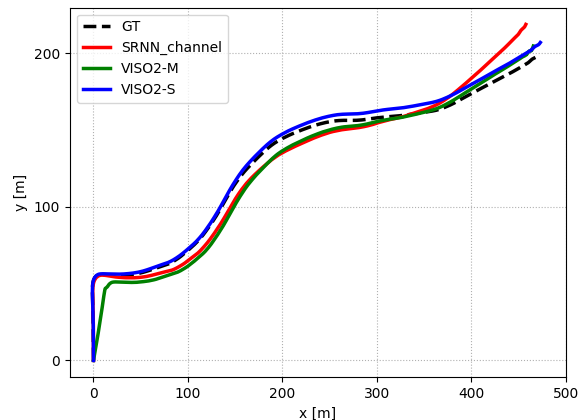}
			\label{fig:trajectory_kitti_00_03}
	\end{minipage}}
	\subfigure[Seq 05]{
		\begin{minipage}{0.48\textwidth}
			\centering
			\includegraphics[height=4.cm]{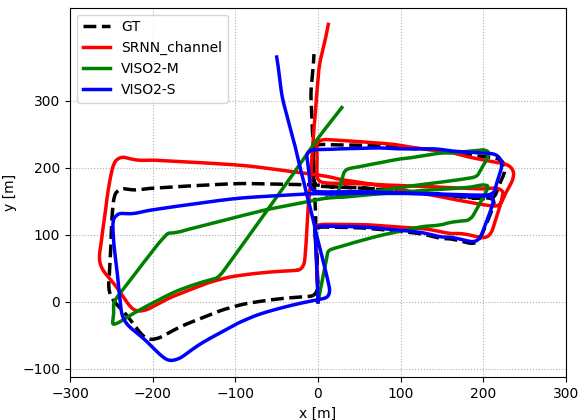}
			\label{fig:trajectory_kitti_00_04}
	\end{minipage}}
	
	\subfigure[Seq 07]{
		\begin{minipage}{0.48\textwidth}
			\centering
			\includegraphics[height=4.cm]{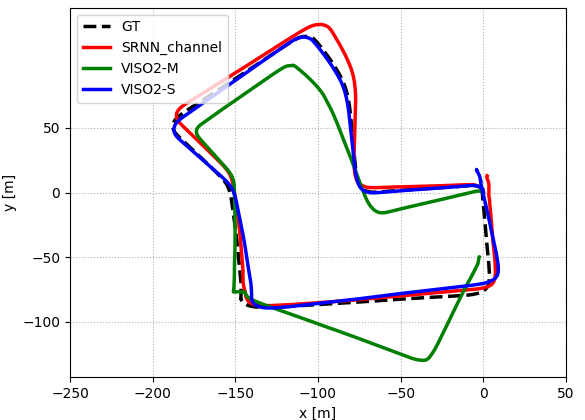}
			\label{fig:trajectory_kitti_00_07}
	\end{minipage}}
	%\qquad
	\subfigure[Seq 10]{
		\begin{minipage}{0.48\textwidth}
			\centering
			\includegraphics[height=4.cm]{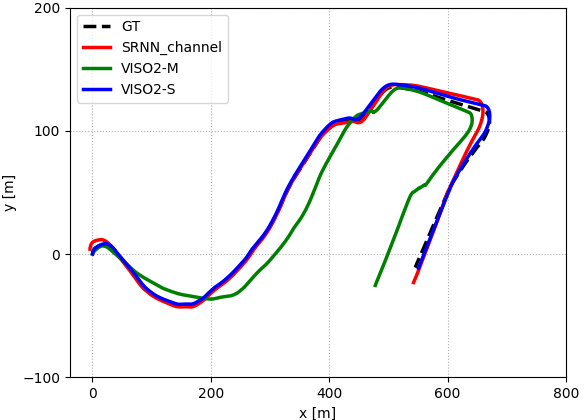}
			\label{fig:trajectory_kitti_00_10_10}
	\end{minipage}}
	\caption{The trajectories of ground-truth, VISO2-M, VISO2-S and our model on Seq 03, 05, 07 and 10 of the KITTI benchmark.}
	\label{fig:trajectory_kitti_00_10}
\end{figure}
\setlength{\tabcolsep}{1.4pt}

%% file: fig_table/table_icl_rpe.tex
%We would be glad to respond to further questions and comments you may have.
%\tiny
%\setlength{\tabcolsep}{2.0pt}
%\setlength\LTleft{-1in}
%\setlength\LTright{-1in plus 1 fill}
\begin{table}[b]
	%\scriptsize
	\centering
		\caption{Evaluation on the ICL\_NUIM dataset. Results of DRFE, DEMO, DVO and MWO are taken directly from \cite{kim2017visual}.}
		\begin{center}
			%\resizebox{\textwidth}{15mm}{
			\begin{tabular}{c|cccccc}
				\hline
				\hline
				Sequence & $\mathbf{SRNN\_channel}$ & $\mathbf{SRNN}$ & DFRE \cite{kim2017visual}& DEMO \cite{demo2014zhang} & DVO \cite{dvo2013robust} & MWO \cite{mwo2016zhou}   \\
				\hline
				lr kt1 & $\mathbf{0.009}$ & 0.010 & 0.021 & 0.020 & 0.023 & 0.100  \\
				lr kt2 & $\mathbf{0.019}$ & 0.019  & 0.031 & 0.090 & 0.084 & 0.052 \\
				\hline
				of kt1 & $\mathbf{0.011}$ & 0.015 & 0.014 & 0.054 & 0.045 & 0.263  \\
				of kt2 & 0.019 & 0.021 & $\mathbf{0.015}$ & 0.079 & 0.065 & 0.047 \\ 
				\hline
				\hline
			\end{tabular}	
		\end{center}
		%\end{left}
	%\end{threeparttable}}
	% Relative pose errors of four algorithms (from \cite{kim2017visual}) and our method on ICL\_NUIM dataset.
	%The four compared algorithms use depth information, while our method utilizes monocular RGB images only.
	%Predicted trajectories of ORB\_SLAM \cite{mur2017orb-slam2} are aligned with ground-truth due to unknown scale. VISO2\_M \cite{geiger2011stereoscan} recovers scale using the fixed camera height as priori.}
	\label{table:icl}
\end{table}
\setlength{\tabcolsep}{1.4pt}

%% file: fig_table/fig_kitti_11_21.tex
\begin{figure}[t]
	\centering
	\subfigure[Seq 11]{
		\begin{minipage}{0.3\textwidth}
			\centering
			\includegraphics[height=2.7cm]{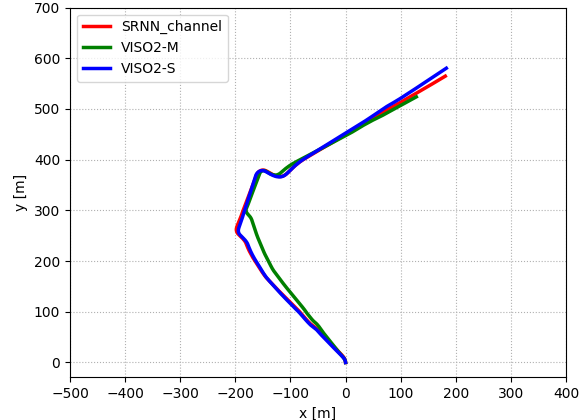}
	\end{minipage}}
	\subfigure[Seq 12]{
		\begin{minipage}{0.3\textwidth}
			\centering
			\includegraphics[height=2.7cm]{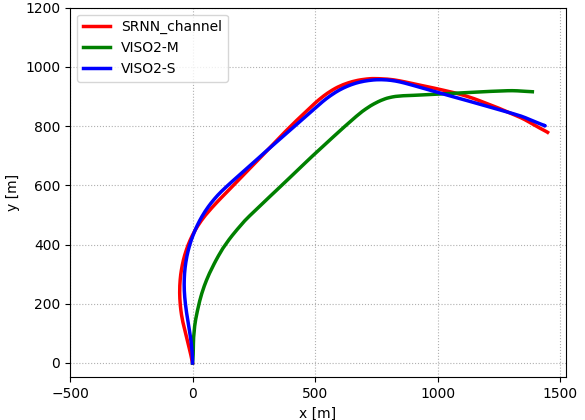}
	\end{minipage}}
	\subfigure[Seq 13]{
		\begin{minipage}{0.3\textwidth}
			\centering
			\includegraphics[height=2.7cm]{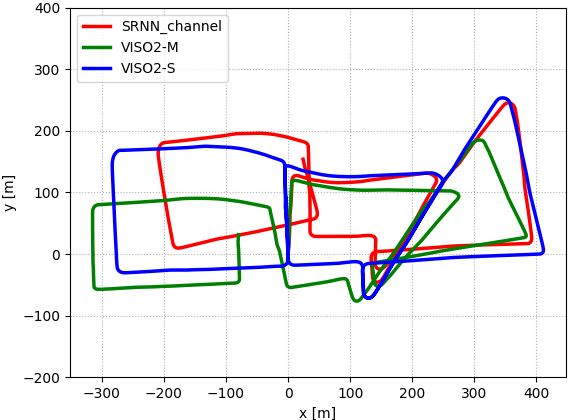}
	\end{minipage}}
	\subfigure[Seq 14]{
		\begin{minipage}{0.3\textwidth}
			\centering
			\includegraphics[height=2.7cm]{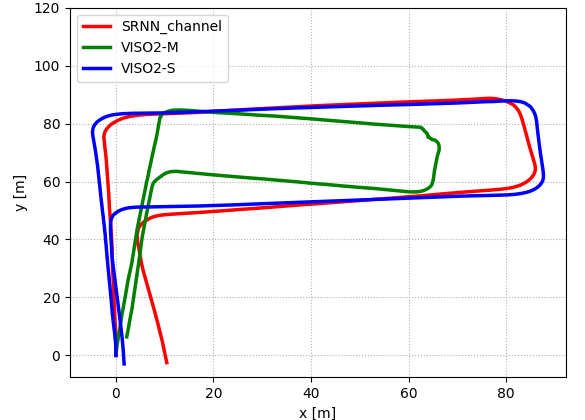}
	\end{minipage}}
	%\vspace{-0.4cm}
	\subfigure[Seq 15]{
		\begin{minipage}{0.3\textwidth}
			\centering
			\includegraphics[height=2.7cm]{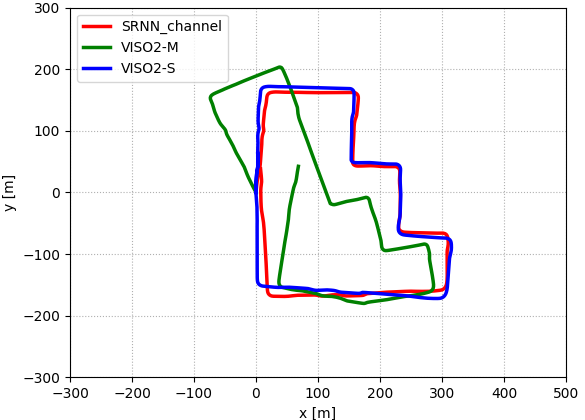}
	\end{minipage}}
	\subfigure[Seq 16]{
		\begin{minipage}{0.3\textwidth}
			\centering
			\includegraphics[height=2.7cm]{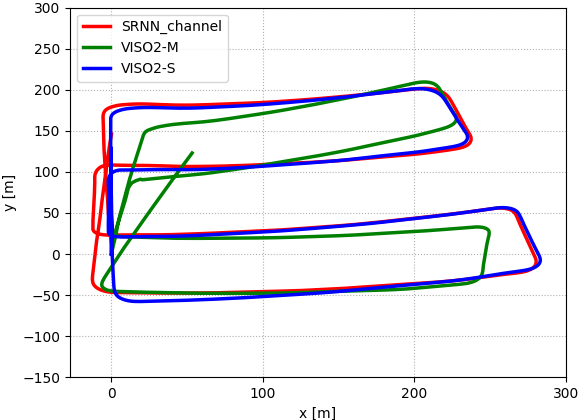}
	\end{minipage}}
	\subfigure[Seq 18]{
		\begin{minipage}{0.3\textwidth}
			\centering
			\includegraphics[height=2.7cm]{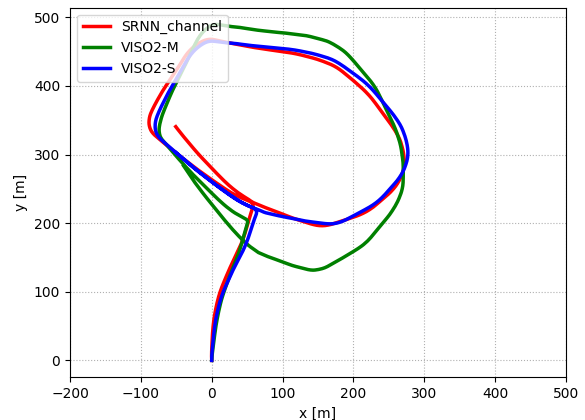}
	\end{minipage}}
	\subfigure[Seq 19]{
		\begin{minipage}{0.3\textwidth}
			\centering
			\includegraphics[height=2.7cm]{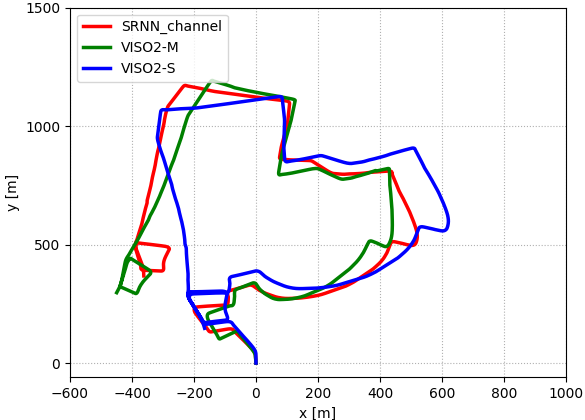}
	\end{minipage}}
	\subfigure[Seq 20]{
		\begin{minipage}{0.3\textwidth}
			\centering
			\includegraphics[height=2.7cm]{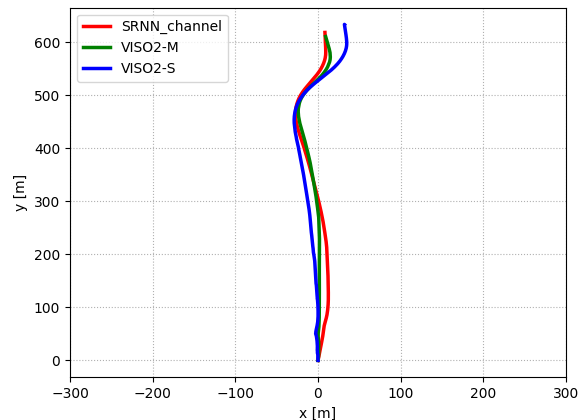}
	\end{minipage}}
	\caption{The predicted trajectories on KITTI sequences 11-20. Results of VISO2-S are used as reference since ground-truth poses of these sequences are unavailable.}
	\label{fig:trajectory_kitti_11-21}
	%\vspace{0.cm}
\end{figure}
\setlength{\tabcolsep}{1.4pt}

%% file: 0082.bbl
\begin{thebibliography}{10}
\providecommand{\url}[1]{\texttt{#1}}
\providecommand{\urlprefix}{URL }
\providecommand{\doi}[1]{https://doi.org/#1}

\bibitem{bazin2010motion}
Bazin, J.C., Demonceaux, C., Vasseur, P., Kweon, I.: {Motion Estimation by
  Decoupling Rotation and Translation in Catadioptric Vision}. CVIU  (2010)

\bibitem{chen2017deeplab}
Chen, L., Papandreou, G., Kokkinos, I., Murphy, K., Yuille, A.: {DeepLab:
  Semantic Image Segmentation with Deep Convolutional Nets, Atrous Convolution,
  and Fully Connected CRFs}. TPAMI  (2018)

\bibitem{choi2018context}
Choi, J., Chang, H.J., Fischer, T., Yun, S., Lee, K., Jeong, J., Demiris, Y.,
  Choi, J.Y.: {Context-aware Deep Feature Compression for High-speed Visual
  Tracking}. In: CVPR (2018)

\bibitem{clark2017vinet}
Clark, R., Wang, S., Wen, H., Markham, A., Trigoni, N.: {VINet: Visual-inertial
  Odometry as A Sequence-to-Sequence Learning Problem}. In: AAAI (2017)

\bibitem{dai2017bundlefusion}
Dai, A., Nie{\ss}ner, M., Zollh{\"o}fer, M., Izadi, S., Theobalt, C.:
  {Bundlefusion: Real-time Globally Consistent 3D Reconstruction Using
  On-the-fly Surface Reintegration}. TOG  (2017)

\bibitem{dosovitskiy2015flownet}
Dosovitskiy, A., Fischer, P., Ilg, E., Hausser, P., Hazirbas, C., Golkov, V.,
  van~der Smagt, P., Cremers, D., Brox, T.: {Flownet: Learning Optical Flow
  with Convolutional Networks}. In: ICCV (2015)

\bibitem{engel2017dso}
Engel, J., Koltun, V., Cremers, D.: {Direct Sparse Odometry}. TPAMI  (2017)

\bibitem{engel2014lsd-slam}
Engel, J., Sch{\"o}ps, T., Cremers, D.: {LSD-SLAM: Large-scale Direct Monocular
  SLAM}. In: ECCV (2014)

\bibitem{geiger2012kitti}
Geiger, A., Lenz, P., Urtasun, R.: {Are We Ready for Autonomous Driving? The
  KITTI Vision Benchmark Suite}. In: CVPR (2012)

\bibitem{geiger2011stereoscan}
Geiger, A., Ziegler, J., Stiller, C.: {Stereoscan: Dense 3D Reconstruction in
  Real-time}. In: IV (2011)

\bibitem{icl2014benchmark}
Handa, A., Whelan, T., McDonald, J., Davison, A.J.: {A Benchmark for RGB-D
  Visual Odometry, 3D Reconstruction and SLAM}. In: ICRA (2014)

\bibitem{he2017mask_rnn}
He, K., Gkioxari, G., Doll{\'a}r, P., Girshick, R.: {Mask R-CNN}. In: ICCV
  (2017)

\bibitem{he2015msra}
He, K., Zhang, X., Ren, S., Sun, J.: {Delving Deep into Rectifiers: Surpassing
  Human-level Performance on Imagenet Classification}. In: ICCV (2015)

\bibitem{hochreiter1997lstm}
Hochreiter, S., Schmidhuber, J.: {Long Short-term Memory}. Neural Computation
  (1997)

\bibitem{hu2018squeeze}
Hu, J., Shen, L., Sun, G.: {Squeeze-and-Excitation Networks}. In: CVPR (2018)

\bibitem{jo2018camera}
Jo, Y., Jang, J., Paik, J.: {Camera Orientation Estimation Using Motion Based
  Vanishing Point Detection for Automatic Driving Assistance System}. In: ICCE
  (2018)

\bibitem{kaess2009flow}
Kaess, M., Ni, K., Dellaert, F.: {Flow Separation for Fast and Robust Stereo
  Odometry}. In: ICRA (2009)

\bibitem{dvo2013robust}
Kerl, C., Sturm, J., Cremers, D.: {Robust Ddometry Estimation for RGB-D
  Cameras}. In: ICRA (2013)

\bibitem{kim2017visual}
Kim, P., Coltin, B., Kim, H.J.: {Visual Odometry with Drift-free Rotation
  Estimation Using Indoor Scene Regularities}. In: BMVC (2017)

\bibitem{Kingma2014Adam}
Kingma, D.P., Ba, J.: {Adam: A Method for Stochastic Optimization}. In: ICLR
  (2015)

\bibitem{lee2015real}
Lee, J.K., Yoon, K.J., et~al.: {Real-time Joint Estimation of Camera
  Orientation and Vanishing Points}. In: CVPR (2015)

\bibitem{li2017undeepvo}
Li, R., Wang, S., Long, Z., Gu, D.: {UnDeepVO: Monocular Visual Odometry
  through Unsupervised Deep Learning}. In: ICRA (2018)

\bibitem{liu2017picanet}
Liu, N., Han, J.: {PiCANet: Learning Pixel-wise Contextual Attention in
  ConvNets and Its Application in Saliency Detection}. In: CVPR (2018)

\bibitem{lowe2004sift}
Lowe, D.G.: {Distinctive Image Features from Scale-invariant Keypoints}. IJCV
  (2004)

\bibitem{mac2018context}
Mac~Aodha, O., Perona, P., et~al.: {Context Embedding Networks}. In: CVPR
  (2018)

\bibitem{mur2017orb-slam2}
Mur-Artal, R., Tard{\'o}s, J.D.: {ORB-SLAM2: An Open-source SLAM System for
  Monocular, Stereo, and RGB-D Cameras}. T-RO  (2017)

\bibitem{relu2010}
Nair, V., Hinton, G.E.: {Rectified Linear Units Improve Restricted Boltzmann
  Machines}. In: ICML (2010)

\bibitem{newcombe2011dtam}
Newcombe, R.A., Lovegrove, S.J., Davison, A.J.: {DTAM: Dense Tracking and
  Mapping in Real-time}. In: ICCV (2011)

\bibitem{pytorch}
Paszke, A., Gross, S., Chintala, S., Chanan, G.: {Pytorch}.
  \url{https://github.com/pytorch/pytorch} (2017)

\bibitem{paz2008large}
Paz, L.M., Pini{\'e}s, P., Tard{\'o}s, J.D., Neira, J.: {Large-scale 6-DOF SLAM
  with Stereo-in-Hand}. T-RO  (2008)

\bibitem{orb2011}
Rublee, E., Rabaud, V., Konolige, K., Bradski, G.: {ORB: An Efficient
  Alternative to SIFT or SURF}. In: ICCV (2011)

\bibitem{straub2015real}
Straub, J., Bhandari, N., Leonard, J.J., Fisher, J.W.: {Real-time Manhattan
  World Rotation Estimation in 3D}. In: IROS (2015)

\bibitem{tardif2008monocular}
Tardif, J.P., Pavlidis, Y., Daniilidis, K.: {Monocular Visual Odometry in Urban
  Environments Using an Omnidirectional Camera}. In: IROS (2008)

\bibitem{ummenhofer2017demon}
Ummenhofer, B., Zhou, H., Uhrig, J., Mayer, N., Ilg, E., Dosovitskiy, A., Brox,
  T.: {DeMoN: Depth and Motion Network for Learning Monocular Stereo}. In: CVPR
  (2017)

\bibitem{wang2017deepvo}
Wang, S., Clark, R., Wen, H., Trigoni, N.: {DeepVO: Towards End-to-end Visual
  Odometry with Deep Recurrent Convolutional Neural Networks}. In: ICRA (2017)

\bibitem{wang2017espvo}
Wang, S., Clark, R., Wen, H., Trigoni, N.: {End-to-end, Sequence-to-sequence
  Probabilistic Visual Odometry through Deep Neural Networks}. IJRR  (2017)

\bibitem{xingjian2015convolutional}
Xingjian, S., Chen, Z., Wang, H., Yeung, D.Y., Wong, W.K., Woo, W.c.:
  {Convolutional LSTM Network: A Machine Learning Approach for Precipitation
  Nowcasting}. In: NIPS (2015)

\bibitem{yin2018geonet}
Yin, Z., Shi, J.: {GeoNet: Unsupervised Learning of Dense Depth, Optical Flow
  and Camera Pose}. In: CVPR (2018)

\bibitem{zamir2018taskonomy}
Zamir, A.R., Sax, A., Shen, W., Guibas, L., Malik, J., Savarese, S.:
  {Taskonomy: Disentangling Task Transfer Learning}. In: CVPR (2018)

\bibitem{zhan2018feature}
Zhan, H., Garg, R., Saroj~Weerasekera, C., Li, K., Agarwal, H., Reid, I.:
  {Unsupervised Learning of Monocular Depth Estimation and Visual Odometry with
  Deep Feature Reconstruction}. In: CVPR (2018)

\bibitem{zhang2018context}
Zhang, H., Dana, K., Shi, J., Zhang, Z., Wang, X., Tyagi, A., Agrawal, A.:
  {Context Encoding for Semantic Segmentation}. In: CVPR (2018)

\bibitem{demo2014zhang}
Zhang, J., Kaess, M., Singh, S.: {Real-time Depth Enhanced Monocular Odometry}.
  In: IROS (2014)

\bibitem{zhou2017egomotion}
Zhou, T., Brown, M., Snavely, N., Lowe, D.G.: {Unsupervised Learning of Depth
  and Ego-motion from Video}. In: CVPR (2017)

\bibitem{mwo2016zhou}
Zhou, Y., Kneip, L., Rodriguez, C., Li, H.: {Divide and Conquer: Efficient
  Density-based Tracking of 3D Sensors in Manhattan Worlds}. In: ACCV (2016)

\end{thebibliography}
